\documentclass{article}
\PassOptionsToPackage{numbers,sort&compress}{natbib}
\usepackage[preprint]{neurips2026/neurips_2026}
\usepackage[utf8]{inputenc}
\usepackage[T1]{fontenc}
\usepackage{microtype}
\usepackage{amsmath,amssymb}
\usepackage{booktabs}
\usepackage{array}
\usepackage{multirow}
\usepackage{enumitem}
\usepackage{xcolor}
\usepackage{graphicx}
\usepackage{xurl} 
\usepackage{pifont}
\usepackage{tikz}
\usetikzlibrary{arrows.meta,positioning,fit,backgrounds,calc}
\usepackage{pgfplots}
\pgfplotsset{compat=1.18}
\usepackage{algorithm}
\usepackage{algpseudocode}
\usepackage{cleveref}
\usepackage{xspace}
\definecolor{NPblue}{HTML}{2B6CB0}
\definecolor{NPteal}{HTML}{2C7A7B}
\definecolor{NPorange}{HTML}{C05621}
\definecolor{NPgray}{HTML}{4A5568}

\newif\ifanon
\newcommand{\arm}[1]{\textsc{#1}}
\newcommand{\code}[1]{\texttt{#1}}
\newcommand{\ctsr}{CTSR}

\anonfalse
\newcommand{\sysname}{Nerveplane\xspace}

\title{Verifying Coordination in Parallel Coding Agents:\\ NP-Bench and a Scheduling Planner}
\author{%
  Sumanyu Muku\\
  Amira Learning\\
  \texttt{mukusumanyu@gmail.com}%
}

\begin{document}
\maketitle
\begin{abstract}
A team of coding agents can look fine agent by agent and still fail as a team: each one passes its
own tests while the merged result is broken, and single-agent evaluation never catches it.
Teams increasingly run several LLM coding agents in parallel on the same codebase, and the
agents get in each other's way: two rewrite the same function, one ships against a contract a
teammate just changed, and integration fails after the work is already done. Most coordination
tools treat this reactively (watch for a conflict, then warn), but under agent-speed concurrency
the warning arrives after the wasted edit. We argue the problem is better posed as
\emph{scheduling}: take each work item's declared scope, partition the work into disjoint scopes,
and order merges along the producer$\rightarrow$consumer dependency graph, all up front. We build
this planner into \sysname, a local-first system for coding agents, and evaluate it
with NP-Bench, an environment-grounded, three-arm benchmark (no coordination; reactive detection;
proactive planning) that verifies integration outcomes off a real git merge, both as a
deterministic simulation and with live agents. Deterministically, the
planner lifts clean-integration success from 1/9 to 9/9 scenarios and drives merge conflicts from
13 to 0; its advantage grows with the number of agents. With live agents on a breaking contract
change, the planner rescues an outcome that both baselines miss on every seed: a clean-integration
rate of 0 under no-coordination and under reactive detection rises to 1.0 with the planner on a
frontier model, and to 0.6 on a small one, while agents respect their assigned scopes (none of 5
strayed). A separate memory result collapses the
repeated-mistake rate from 1.00 to 0.00 on both a strong and a weak model: a decision recorded in
one session stops an agent in a later session from repeating it, an information effect no
amount of model strength substitutes for. We also report what did \emph{not} help: at
window-fitting scales, routing facts to agents does not rescue retrieval accuracy over long
contexts; its value is cost and capacity, not attention. Methodologically, NP-Bench doubles as an
environment-grounded \emph{verifier} for multi-agent coding: each agent passes its own checks, yet the
integrated system can still be broken, and a real merge-and-test reveals it. Across the two capability
tiers and two vendors we tested, the benefit did not shrink as models got stronger, because it comes
from how work is allocated rather than from model reasoning.
\end{abstract}

\section{Introduction}

Running one coding agent is now routine; running several at once is where teams are headed.
Worktree multiplexers spin up an agent per branch, CI pipelines fan issues out to a pool of
agents, and a single developer can supervise a handful of them at a time. Once that happens, the
bottleneck stops being how well any one agent writes code and starts being whether the agents step
on each other. Two agents edit the same file and produce a merge conflict; one agent renames a
field in an API while another codes against the old shape and integrates to something that merges
cleanly but is broken; a decision made in the morning is re-litigated in the afternoon by an agent
that never saw it. These are coordination failures, and they scale with the number of agents.

The prevailing response is reactive. A coordination layer watches for trouble (overlapping
edits, a changed contract) and warns the affected agent once the change is observed. This is the
same idea behind decades of developer ``workspace awareness'' tools \citep{brun2011proactive}, and
it is genuinely useful for humans who work on the scale of hours. But agents work on the scale of
seconds. By the time a teammate's edit has been sensed and a warning routed, the warned agent has
often already written the conflicting code. The warning then buys a cleanup, not a prevention.

This paper takes a different stance: \textbf{coordination among parallel coding agents is a
scheduling problem}. Before the agents start, take each work item's declared \emph{scope}, partition
the work so that concurrently-running agents have disjoint scopes, and order the merges along the
producer$\rightarrow$consumer dependency graph so that whoever changes a contract lands before
whoever consumes it. Where a reactive detector can only tell an agent it has a problem after the
fact, a schedule prevents the problem by construction.

There is an obvious objection: ``disjoint scopes don't conflict'' is a tautology. But that
objection is the research question in disguise. A schedule helps only if (a) its input scopes are
accurate and (b) the agents actually \emph{respect} the scopes they are handed. In this paper the
scopes are \emph{declared} (given by each work item's stated files and consumed contracts), so we
assume (a) and measure (b): an agent told to stay out of a file may edit it anyway, and whether it
complies is exactly what the live experiments check. Inferring scopes automatically is an open
problem we flag but do not solve. The contribution is the compliance measurement, not the tautology.

There is a verification problem hiding in that measurement. Whether a coordination layer actually
helps is not something an agent can self-report, and it is invisible to single-agent evaluation: each
agent, checked on its own branch, looks correct; the breakage surfaces only when the branches are
integrated. Answering it takes an \emph{environment-grounded} verifier that reads ground-truth
signals (does the merged tree build, does it pass, did the consumer adapt to the new
contract) off a real integration rather than trusting any agent's word. NP-Bench is that verifier,
and its three-arm design turns ``did coordination help'' from an assertion into a measurement.

We make four claims and back each with data.
\begin{enumerate}[leftmargin=1.4em,itemsep=1pt,topsep=2pt]
\item \textbf{A verification benchmark.} NP-Bench scores three arms (no coordination \arm{C0}, reactive
detection \arm{C1-detect}, proactive planning \arm{C1-plan}) both deterministically and with live
agents from two model families, reading ground-truth integration signals off a real merge, the
coordination axis that single-agent benchmarks like SWE-bench do not measure (\Cref{sec:npbench}). We
present it as a focused study of \emph{when} coordination helps, not a comprehensive benchmark.
\item \textbf{A formulation and a planner.} We cast coordination as disjoint-scope partitioning plus
topological merge ordering, and implement a small, deterministic static planner
(\Cref{sec:planner}).
\item \textbf{Results with an honest capability analysis.} In the evaluated scenarios, the planner
strictly dominates both baselines; a vague reactive warning is worth about as much as nothing to a
live agent; and we separate the part of the effect that holds across the tested capability tiers
from the part that varies with capability
(\Cref{sec:results,sec:discussion}).
\item \textbf{What did not work.} We tried to show that routing facts to agents rescues retrieval
accuracy over long contexts and could not; we report that negative result and the measurement bug we
caught along the way (\Cref{sec:negatives}).
\end{enumerate}

\section{Related work}
\label{sec:related}

\paragraph{Multi-agent LLM frameworks.} A first wave of systems orchestrates several LLM agents
through roles and messages: AutoGen composes conversable agents under a group-chat manager
\citep{wu2023autogen}; MetaGPT runs an SOP-driven ``software company'' of agents
\citep{hong2024metagpt}; ChatDev structures a chat-driven development organization
\citep{qian2024chatdev}; CAMEL and AgentVerse study role-play and emergent multi-agent behavior
\citep{li2023camel,chen2023agentverse}. These divide labor by role and coordinate through dialogue.
None offers a scheduling guarantee that concurrently-editing agents will avoid colliding on shared
files or stale contracts. The planner here is complementary: it takes the work items such a
framework produces and computes a disjoint-scope partition and merge order before execution. We
target the substrate (who may touch what, and in what order), not the role or dialogue layer.

\paragraph{Coding-agent benchmarks.} Evaluation of coding agents is dominated by single-agent,
single-repository task success: SWE-bench resolves real GitHub issues \citep{jimenez2024swebench},
SWE-agent studies the agent-computer interface that drives such resolution
\citep{yang2024sweagent}, and execution benchmarks like HumanEval \citep{chen2021codex} and
LiveCodeBench \citep{jain2024livecodebench} score generated code. None measures whether $N$ agents
working in parallel integrate cleanly. NP-Bench adds exactly that axis (clean-merge success,
wasted work, and scope compliance for \emph{concurrent} agents) and is orthogonal to per-task
resolution.

\paragraph{Conflict detection and developer coordination.} Proactively predicting merge conflicts
by watching parallel edits is a long line in software engineering
\citep{brun2011proactive}. This is the intellectual ancestor of the \arm{C1-detect} arm: valuable,
but reactive: the warning fires once a change is observed, which under agent-speed concurrency is
frequently after the wasted edit. We show empirically that a proactive schedule dominates reactive
detection, and that a vague reactive warning is close to no coordination for a live agent.

\paragraph{Scheduling and dependency-aware ordering.} Framing coordination as scheduling connects
to classic task-graph work: topological ordering of a dependency DAG \citep{kahn1962topological}
and the NP-hardness of the partitioning/coloring subproblem \citep{garey1979computers}. That
software structure mirrors team structure is Conway's observation \citep{conway1968committees}, and
that adding people (or agents) to concurrent work raises coordination cost is Brooks's
\citep{brooks1975mythical}. We adapt disjoint-partition-plus-topological-merge to the
\emph{declared code scopes} of LLM agents, where the novel empirical question studied here is
whether agents will respect the resulting assignments. Inferring those scopes automatically remains
future work.

\paragraph{Agent protocols and memory.} Agent tooling is standardizing on protocols for
model-to-tool and agent-to-agent communication---the Model Context Protocol
\citep{anthropic2024mcp} and Agent2Agent \citep{google2025a2a}. \sysname exposes its coordination
surface over MCP but adds the missing piece those protocols leave open: what to route, to whom, and
in what order. On memory, systems such as MemGPT \citep{packer2023memgpt} and generative agents
\citep{park2023generative} give an agent long-term recall within its own task. Our memory result is
cross-session and cross-agent: a decision recorded by one session prevents an agent in a later
session from repeating a mistake, and we isolate it as an information effect that holds
equally on strong and weak models.

\Cref{tab:related} places these lines side by side.

\begin{table}[t]
\centering\small
\caption{Where NP-Bench and the planner sit relative to adjacent work.}
\label{tab:related}
\resizebox{\linewidth}{!}{%
\begin{tabular}{@{}lcccc@{}}
\toprule
System / benchmark & Multi-agent & Coord.\ guarantee & Proactive & Parallel integration \\
\midrule
AutoGen / MetaGPT / ChatDev / CAMEL & yes (roles) & no (emergent) & --- & no \\
SWE-bench, HumanEval, LiveCodeBench & no (single agent) & n/a & n/a & no \\
Workspace-awareness / conflict warn & yes (humans) & no (warn only) & no (reactive) & partial \\
Build/merge DAG systems & n/a & yes (deps) & yes & no (not agents) \\
Agent memory (MemGPT, gen.\ agents) & partial & n/a & n/a & no \\
\textbf{NP-Bench + planner (this work)} & \textbf{yes} & \textbf{yes} & \textbf{yes} & \textbf{yes} \\
\bottomrule
\end{tabular}}
\end{table}

\section{The coordination plane}
\label{sec:system}

The planner does not run in a vacuum; it needs to know what agents exist, what they are touching,
and how services depend on one another. \sysname supplies that substrate. It is a local-first
system for coding agents: a single per-machine daemon, a SQLite store, and a set of
tools exposed over MCP (\code{register}, \code{sync}, \code{publish}, \code{decision},
\code{discover}, \code{chat}, \code{memory}, and \code{task}); the planner runs inside the daemon
and delivers each agent's assignment through \code{register}/\code{sync}. It is
agent-agnostic: Claude Code, Codex, and OpenCode all connect through the same surface.

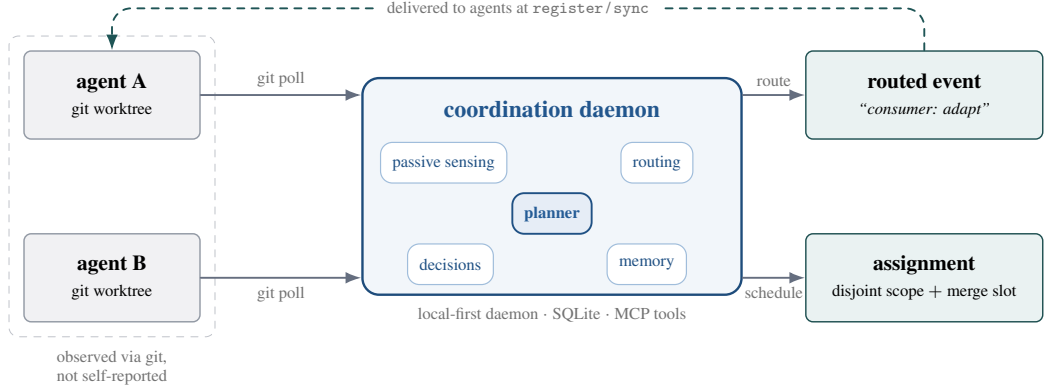
\begin{figure}[t]
\centering
\resizebox{0.98\linewidth}{!}{%
\begin{tikzpicture}[
  font=\small, >=Latex, line join=round, line cap=round,
  agent/.style={draw=NPgray!60, line width=0.6pt, rounded corners=3pt, fill=NPgray!8,
                align=center, minimum width=26mm, minimum height=13mm, inner sep=4pt},
  daemon/.style={draw=NPblue!70!black, line width=1pt, rounded corners=7pt, fill=NPblue!6,
                 minimum width=56mm, minimum height=32mm},
  pill/.style={draw=NPblue!45, line width=0.5pt, rounded corners=5pt, fill=white,
               font=\scriptsize, text=NPblue!75!black, inner xsep=5pt, inner ysep=3pt,
               minimum height=6mm},
  plan/.style={pill, draw=NPblue!75!black, line width=0.8pt, fill=NPblue!12,
               text=NPblue!85!black, font=\scriptsize\bfseries},
  obox/.style={draw=NPteal!65!black, line width=0.6pt, rounded corners=3pt, fill=NPteal!10,
               align=center, minimum width=35mm, minimum height=13mm, inner sep=4pt},
  flow/.style={->, draw=NPgray!85, line width=0.8pt},
  back/.style={->, draw=NPteal!70!black, line width=0.8pt, dashed},
  lbl/.style={font=\scriptsize, text=black!55},
]
\node[agent] (a) at (0,1.35) {\textbf{agent A}\\[1pt]\scriptsize git worktree};
\node[agent] (b) at (0,-1.35) {\textbf{agent B}\\[1pt]\scriptsize git worktree};
\node[draw=NPgray!35, dashed, rounded corners=4pt, fit=(a)(b), inner sep=6pt] (agents) {};
\node[lbl, below=2pt of agents, align=center] {observed via git,\\not self-reported};
\node[daemon] (d) at (6.5,0) {};
\node[font=\bfseries, text=NPblue!80!black, anchor=north] at ([yshift=-2.2mm]d.north)
  {coordination daemon};
\node[pill] (p1) at (4.9,0.35)  {passive sensing};
\node[pill] (p2) at (8.05,0.35) {routing};
\node[plan] (p3) at (6.5,-0.4)  {planner};
\node[pill] (p4) at (5.0,-1.15) {decisions};
\node[pill] (p5) at (7.9,-1.15) {memory};
\node[lbl, below=1.5pt of d] {local-first daemon $\cdot$ SQLite $\cdot$ MCP tools};
\node[obox] (r) at (12.0,1.35)  {\textbf{routed event}\\[1pt]\scriptsize\itshape ``consumer: adapt''};
\node[obox] (o) at (12.0,-1.35) {\textbf{assignment}\\[1pt]\scriptsize disjoint scope $+$ merge slot};
\draw[flow] (a.east) -- node[above, lbl]{git poll} (d.west|-a);
\draw[flow] (b.east) -- node[below, lbl]{git poll} (d.west|-b);
\draw[flow] (d.east|-r) -- node[above, lbl]{route} (r.west);
\draw[flow] (d.east|-o) -- node[below, lbl]{schedule} (o.west);
\draw[back, rounded corners=8pt] (r.north) -- ++(0,0.62) -| (a.north);
\node[lbl, fill=white, inner xsep=5pt, inner ysep=1.5pt] at (6.0,2.62)
  {delivered to agents at \code{register}\,/\,\code{sync}};
\end{tikzpicture}}
\caption{\sysname observes each agent's worktree by polling git (\emph{passive sensing}), then
routes only the relevant events and, with the planner, hands each agent a disjoint scope and a
merge slot \emph{before} the edits happen, rather than warning after a collision is sensed.}
\label{fig:arch}
\end{figure}

The design choice that matters for coordination is that the daemon observes state rather than
waiting to be told about it. A \emph{passive sensing} loop polls each registered agent's git
worktree and emits an event when the changed-file set actually changes, so coordination never
depends on an agent remembering to announce its work.\footnote{\ifanon
\else
The system and the NP-Bench harness are open source (\sysname v0.17.0).\fi} A routing engine then decides who
needs to hear about a given change: the owner of an affected task, other agents in the
same repository, and, through a service graph, agents in \emph{other} repositories that consume a
changed API contract. Durable project decisions live in a queryable ledger separate from chat
history, and a cross-agent memory (keyword by default, optionally semantic) lets a fact learned in
one session resurface in another. \Cref{fig:arch} sketches the loop.

The contrast with a message bus is deliberate. A bus moves whatever agents choose to send; a
state-aware plane understands branches, changed files, and service contracts, and can therefore
route by relevance and, with the planner, act \emph{before} the edits happen rather than after.

\section{Coordination as scheduling}
\label{sec:planner}

\paragraph{Model.} Let $W$ be a set of work items. Each item $w$ has a \emph{declared scope}
$\sigma(w)$---the set of files (or symbols) it will write---and a set of contracts it
\emph{consumes}. Two items \emph{conflict} when their scopes intersect,
$\sigma(w_i)\cap\sigma(w_j)\neq\varnothing$. Item $w_j$ \emph{depends on} $w_i$ (edge
$w_i\!\rightarrow\!w_j$) when $w_j$ consumes a contract that $w_i$ produces, i.e.\ $w_i$'s scope
overlaps $w_j$'s consumed set. A \emph{plan} is a partition of $W$ into batches whose members have
pairwise-disjoint scopes, together with a merge order $\pi$ that respects the dependency edges.

\paragraph{Claim (assumption-scoped).} If the declared scopes are accurate and the dependency
relation is a DAG, then a disjoint-scope partition plus a topological merge order yields
\emph{zero} structural conflicts by construction: concurrently-running items never touch the same
file, and no consumer is merged before its producer. A reactive detector cannot make this guarantee
because it acts only after a collision has been sensed. Scope accuracy is assumed through the
hand-declared inputs; agent scope-compliance is the empirical unknown measured in
\Cref{sec:results}.

\paragraph{Algorithm.} \Cref{alg:plan} is the static planner. It
builds the dependency graph, marks items that must not take their naive edit (either because they
contend for an owned file or because they consume a produced contract), computes a stable
topological order by Kahn's algorithm \citep{kahn1962topological} with deterministic cycle
breaking, and greedily colors the scope-overlap graph into parallel batches. Exact minimum coloring
is NP-hard \citep{garey1979computers}; the greedy pass is a documented approximation and is enough
for the suite. Every tie breaks on input order, so a given input always yields the same plan, a
requirement for reproducible experiments. The item's assignment (its scope, merge slot, and a
human-readable reason when it was reassigned) is delivered to the agent at \code{register}/\code{sync}
time.

\begin{algorithm}[t]
\caption{\textsc{BuildPlan} (static coordination planner)}
\label{alg:plan}
\begin{algorithmic}[1]
\Require work items $W$, each with scope $\sigma(w)$ and consumed set $c(w)$
\Ensure batches (disjoint-scope), merge order $\pi$, reassigned set $R$
\State build edges $w_i\!\rightarrow\!w_j$ for all $i\neq j$ with $\sigma(w_i)\cap c(w_j)\neq\varnothing$
\State $R \gets \varnothing$; for each file, its first writer (input order) is owner
\ForAll{items $w$ touching a file owned by another item}
    \State add $w$ to $R$ \Comment{scope contention}
\EndFor
\ForAll{items $w$ with $\geq 1$ producer}
    \State add $w$ to $R$ \Comment{must adapt to a consumed contract}
\EndFor
\State $\pi \gets$ \textsc{Kahn}(edges) with input-order tie-break; break cycles by dropping the edge from the producer with fewest downstream consumers
\State greedily assign each item (in order $\pi$) to the lowest batch with no scope-overlapping item
\State \Return batches, $\pi$, $R$
\end{algorithmic}
\end{algorithm}

The reassigned set $R$ is what the suite ultimately probes. An item lands in $R$ when the planner
decides, up front and independent of runtime timing, that its naive scope would collide with a
teammate or that it consumes a contract another item produces. That is exactly the information a
reactive detector can only deliver after the agent has already edited.

\section{NP-Bench}
\label{sec:npbench}

NP-Bench compares three arms. \arm{C0} is no coordination: agents act blind to one another.
\arm{C1-detect} is reactive coordination: an agent is warned when a teammate's sensed change touches
a file it cares about, and (for live agents) the warning is deliberately vague (``a teammate may
be editing $X$''), because that is what a realistic just-in-time warning looks like. \arm{C1-plan}
is the proactive planner: before starting, an agent receives either an allowed/forbidden scope or,
for a contract consumer, the exact contract migration and an instruction to code against the new
shape after the producer merges. The plan also supplies the merge order. Thus the live comparison
evaluates the combined effect of timing, information specificity, scope assignment, and merge
ordering rather than isolating any one component.

The suite runs at two tiers. \textbf{Tier A} is a deterministic simulation that drives the real
product primitives---agent registration, worktree sensing, conflict detection, and the
planner core---over scripted scenarios, and scores the result with a real git
integration. An agent takes its coordinated edit exactly when it was warned in time (\arm{C1-detect})
or reassigned by the planner (\arm{C1-plan}); otherwise it takes its naive edit. Tier A is
CI-gated and reproducible. \textbf{Tier B} replaces the simulated agents with live ones: real
coding agents run headless in isolated git worktrees under each arm, their branches are integrated
with a real merge, and the outcome is scored. Tier B is nondeterministic and costs money, so it is
not a CI gate.

We report the following metrics. \emph{\ctsr{}} (clean-tree success rate) is the fraction of runs
that both merge without conflict and pass a post-integration acceptance check. \emph{Merge
conflicts} and \emph{wasted LOC} count git conflicts and the lines in conflicting edits that must
be reworked. \emph{Scope-leakage} is the fraction of reassigned \arm{C1-plan} agents that edited a
file they were told not to; 0 means full compliance. \emph{Consumer-adaptation rate} is the
fraction of fan-out consumers whose merged code uses the new contract field. \emph{Repeated-mistake
rate} is one minus adherence to a prior session's recorded decision. \emph{Merge-order correctness}
checks that every producer is ordered before its consumers.

Live runs span two model families: two Claude models of very different capability---a frontier model
(Claude Opus~4.8, \code{claude-opus-4-8}) and a small one (Claude Haiku~4.5,
\code{claude-haiku-4-5})---and OpenAI's Codex (\code{gpt-5-codex}). The two Claude models separate
what coordination does from what raw capability does; the second vendor's model checks that the
effect is not a Claude-specific quirk. Live sample sizes are modest ($K\!=\!3$--$15$ seeds per cell, the pivotal contract cell at
$K{=}15$); we report Wilson 95\% intervals and treat the smaller cells as pilots (\Cref{sec:threats}).

\section{Results}
\label{sec:results}

\Cref{tab:headline} consolidates the headline numbers; the rest of this section reads them off one
experiment at a time.

\begin{table}[t]
\centering\small
\caption{NP-Bench headline. \arm{C1-plan} dominates both baselines; \arm{C1-detect}$\approx$\arm{C0}
on every live metric. Tier A is deterministic; Tier B is live agents.}
\label{tab:headline}
\begin{tabular}{@{}llcccc@{}}
\toprule
\# & Experiment (metric) & \arm{C0} & \arm{C1-detect} & \arm{C1-plan} & Tier \\
\midrule
1 & Deterministic 3-arm, 9 scenarios (\ctsr{}) & 1/9 & 4/9 & \textbf{9/9} & A \\
1 & \quad merge conflicts (total) & 13 & 12 & \textbf{0} & A \\
1 & \quad wasted LOC (total) & 54 & 49 & \textbf{0} & A \\
2 & Effect-by-$N$ contention, $N{=}8$ (wasted LOC) & 28 & 28 & \textbf{0} & A \\
3 & Live contract migration, Opus 4.8 (\ctsr{}) & 0.00 & 0.00 & \textbf{1.00} & B \\
3 & Live contract migration, Haiku 4.5 (\ctsr{}) & 0.00 & 0.00 & \textbf{0.60} & B \\
3 & Live contract migration, GPT-5-Codex (\ctsr{}) & 0.00 & 0.00 & \textbf{1.00} & B \\
4 & Live fan-out, $N{\in}\{2,4,8\}$ (adaptation) & 0.00 & 0.00 & \textbf{1.00} & B \\
5 & Same-file assignment, Opus 4.8 (scope-leakage) & --- & --- & \textbf{0.00} & B \\
6 & Memory continuity, both models (repeated-mistake) & 1.00 & --- & \textbf{0.00} & B \\
\bottomrule
\end{tabular}
\end{table}

\subsection{Deterministic three-arm and effect-by-$N$}

Across nine scripted scenarios (shared-file, contract, cross-repo fan-out, an independent control,
and concurrent/high-$N$ variants), \ctsr{} rises from 1/9 under \arm{C0} to 4/9 under
\arm{C1-detect} to 9/9 under \arm{C1-plan}; total merge conflicts fall 13$\rightarrow$12$\rightarrow$0
and wasted LOC 54$\rightarrow$49$\rightarrow$0. The
gap between the two coordination arms is the point. On \emph{sequential} dependency scenarios the
reactive detector and the planner tie: both fix the problem, because the warning has time to
land. On \emph{concurrent} scenarios the detector collapses to the \arm{C0} baseline: the warning
arrives after the edit. The planner does not, because it never let the collision be scheduled in
the first place. On the control scenario (unrelated tasks) all three arms succeed and the plane
raises no warnings, so the coordination layer does no harm when there is nothing to coordinate.
Cross-repo routing is exact: on the fan-out scenario the two real consumers are warned and the
unrelated service is not (100\% routing hit, 0\% false positives).

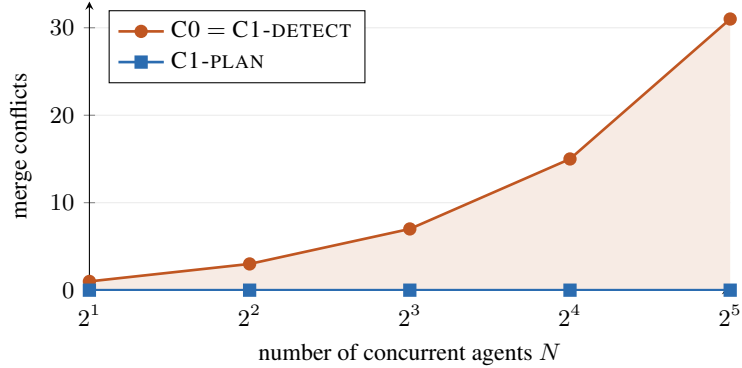
\begin{figure}[t]
\centering
\begin{tikzpicture}
\begin{axis}[
  width=0.72\linewidth, height=5.4cm,
  xlabel={number of concurrent agents $N$}, ylabel={merge conflicts},
  xtick={2,4,8,16,32}, xmode=log, log basis x=2, ymin=0, ymax=33,
  legend pos=north west, legend cell align=left, font=\small,
  ymajorgrids, grid style={black!8}, tick align=outside, axis lines=left,
  every axis plot/.append style={line width=1pt},
]
\addplot[draw=none, fill=NPorange!12, forget plot] coordinates {(2,1)(4,3)(8,7)(16,15)(32,31)} \closedcycle;
\addplot[NPorange, mark=*, mark options={fill=NPorange}] coordinates {(2,1)(4,3)(8,7)(16,15)(32,31)};
\addlegendentry{\arm{C0} $=$ \arm{C1-detect}}
\addplot[NPblue, mark=square*, mark options={fill=NPblue}] coordinates {(2,0)(4,0)(8,0)(16,0)(32,0)};
\addlegendentry{\arm{C1-plan}}
\end{axis}
\end{tikzpicture}
\caption{Deterministic effect-by-$N$ (contention on one file). \arm{C1-detect} tracks \arm{C0}
because the reactive warning is always too late under concurrency; \arm{C1-plan} holds at zero
conflicts at every scale. Wasted LOC follows the same shape at $\approx 4(n{-}1)$.}
\label{fig:byn}
\end{figure}

The effect grows with the number of agents. \Cref{fig:byn} sweeps $N$ agents concurrently rewriting
one hot file. Conflicts scale as $n{-}1$ and wasted LOC as $\approx 4(n{-}1)$ for both \arm{C0} and
\arm{C1-detect}; the planner holds at zero across $N\in\{2,4,8,16,32\}$. The reactive detector
tracks no-coordination exactly here---under concurrency its warning is always too late---so the
advantage of planning is monotone in $N$.

\subsection{Live agents: the planner lifts clean-integration outcomes}

Deterministic results can be dismissed as a simulation we built, so the live runs are the main
evidence. In the contract-migration scenario, a producer renames an invoice money
field (\code{amount} in dollars $\rightarrow$ \code{amountCents} in integer cents); the consumer,
working from the old contract in its own worktree, implements a total. The two files merge cleanly,
but the consumer is semantically broken unless it adapts. \Cref{tab:contract} gives \ctsr{} for
three models across two vendors (the pivotal frontier cell at $K{=}15$, the others at $K{=}5$) with
Wilson 95\% confidence intervals on the planner arm.

\begin{table}[t]
\centering\small
\caption{Live contract-migration \ctsr{} across two model families, with Wilson 95\% CIs on the
planner arm. The result holds across the tested capability tiers and two vendors (all three models
fail at 0.00 without the signal); acting on the plan varies with capability (the two frontier models
$1.00$, the small model $0.60$). On the frontier model the planner and baselines separate with non-overlapping
intervals (15/15 vs.\ 0/15; Fisher's exact $p<10^{-4}$).}
\label{tab:contract}
\begin{tabular}{@{}lccccc@{}}
\toprule
model & $K$ & \arm{C0} & \arm{C1-detect} & \arm{C1-plan} & \arm{C1-plan} 95\% CI \\
\midrule
Claude Opus 4.8 \small(\code{claude-opus-4-8}) & 15 & 0.00 & 0.00 & \textbf{1.00} & $[0.80, 1.00]$ \\
Claude Haiku 4.5 \small(\code{claude-haiku-4-5}) & 5 & 0.00 & 0.00 & \textbf{0.60} & $[0.23, 0.88]$ \\
OpenAI GPT-5-Codex \small(\code{gpt-5-codex}) & 5 & 0.00 & 0.00 & \textbf{1.00} & $[0.57, 1.00]$ \\
\bottomrule
\end{tabular}
\end{table}

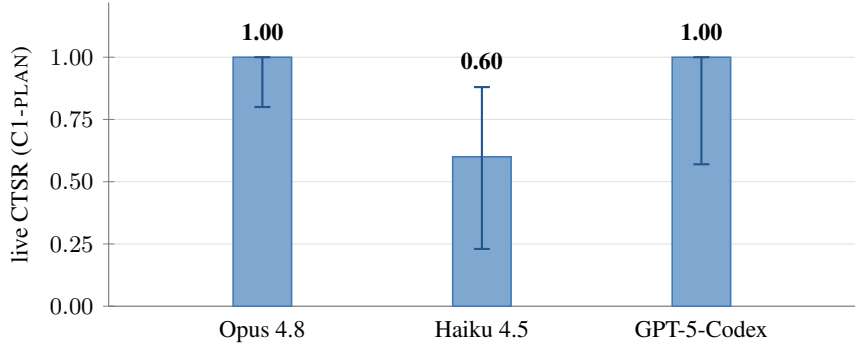
\begin{figure}[t]
\centering
\begin{tikzpicture}
\begin{axis}[
  ybar, bar width=22pt, width=0.82\linewidth, height=5.6cm,
  ymin=0, ymax=1.22, ytick={0,0.25,0.5,0.75,1},
  scaled y ticks=false,
  yticklabel style={/pgf/number format/fixed, /pgf/number format/fixed zerofill,
    /pgf/number format/precision=2, font=\small},
  ylabel={live \ctsr{} (\arm{C1-plan})},
  ylabel style={font=\small},
  symbolic x coords={Opus 4.8, Haiku 4.5, GPT-5-Codex}, xtick=data,
  x tick label style={font=\small},
  ymajorgrids, grid style={black!12},
  tick align=outside, xtick style={draw=none},
  axis lines=left, axis line style={draw=black!55, -},
  clip=false,
  error bars/error bar style={NPblue!80!black, line width=0.8pt},
  error bars/error mark options={rotate=90, mark size=3pt, line width=0.8pt},
  enlarge x limits=0.35,
]
\addplot+[ybar, fill=NPblue!60, draw=NPblue!90, line width=0.6pt,
  error bars/.cd, y dir=both, y explicit]
  table[row sep=\\, x=model, y=val, y error plus=ep, y error minus=em] {
    model val ep em \\
    {Opus 4.8}    1.0 0.0  0.20 \\
    {Haiku 4.5}   0.6 0.28 0.37 \\
    {GPT-5-Codex} 1.0 0.0  0.43 \\
  };
\node[font=\small\bfseries, anchor=south, yshift=3pt] at (axis cs:{Opus 4.8},1.0) {1.00};
\node[font=\small\bfseries, anchor=south, yshift=3pt] at (axis cs:{Haiku 4.5},0.88) {0.60};
\node[font=\small\bfseries, anchor=south, yshift=3pt] at (axis cs:{GPT-5-Codex},1.0) {1.00};
\end{axis}
\end{tikzpicture}
\caption{Live contract-migration \ctsr{} of the planner (\arm{C1-plan}) with Wilson 95\% CIs, by
model (Claude Opus~4.8, Claude Haiku~4.5, OpenAI GPT-5-Codex); both baselines (\arm{C0},
\arm{C1-detect}) sit at the floor, $0.00$, for every model. Opus~4.8 is $K{=}15$, the others
$K{=}5$.}
\label{fig:live}
\end{figure}

The planner lands the breaking change where both baselines fail on
every seed (the frontier models) or most seeds (Haiku). \arm{C1-detect} does no better than \arm{C0},
sitting at 0.00 on every model: the raw replies show the consumer knew ``something may change'' and still coded
to the stale field. A heads-up without the specifics and a merge order buys nothing; only the
explicit up-front assignment carries the contract shape the agent needs. We read this as the
\emph{combined} planner intervention (up-front timing together with the exact scope and
contract) beating a vague, late warning. The live setting does not isolate timing from information
specificity, and the detailed-but-late ablation that would (\Cref{sec:threats}) is future work. The failure of \arm{C0} is not a capability failure:
the information the consumer needs is simply not in its worktree, so no amount of model strength
recovers it, and the same pattern holds across a second vendor's model, so it is not a
Claude-specific artifact.

The fan-out version scales that scenario: one producer migrates the contract and $N$ consumers each
render in their own file. Adaptation rate is 0.00 under both baselines and 1.00 under the planner at
every $N\in\{2,4,8\}$ (frontier, $K{=}3$). The rate saturates, so the absolute number of broken
consumers under \arm{C0} grows linearly with $N$ while the planner holds all of them.

Finally, the make-or-break question for the whole approach: do agents respect the scopes they are
assigned? On the shared-file scenario ($K{=}5$, frontier), scope-leakage is 0.00: every planner
agent put its new work in a separate file and left the owned file untouched, several of them noting
that they were doing so ``as coordinated.'' The disjoint-scope partition is not a tautology defeated
by non-compliance; real agents honor it.

\subsection{Memory continuity: a capability-independent effect}

The results above concern work allocation. A separate experiment concerns knowledge. No model,
however strong, can recall a project-specific fact it was never told; if one session discovers a
gotcha and records it, an agent in a later session should stop repeating the mistake---but
only if memory carries the discovery across sessions. We built a sandbox repository with two
symmetric authentication backends that both export the same function, where which one is sanctioned
appears nowhere in the code; it is a team decision that lives only in memory. Session~1 records the
decision; session~2 (a fresh session) is asked which backend to import. The repeated-mistake rate
falls from 1.00 to 0.00 on both the frontier model and Haiku ($K{=}5$). Because the correct choice
is not recoverable from the repository, the gap is an information gap, not a reasoning gap, and it
closes identically regardless of model strength. The raw replies confirm it: without memory the
agents call the two backends ``a genuine tie'' and guess; with memory they cite the recalled policy
and import the sanctioned module.

\section{What did not work}
\label{sec:negatives}

We initially hypothesized a second result: that routing a needed fact to an agent would rescue
its \emph{accuracy} over a long, distracting context, an ``attention rescue'' to complement the
lost-in-the-middle literature \citep{liu2024lost}. It did not hold. We report it anyway, because it
changes what one should claim for context routing.

At the modest scales we first tested, a frontier model recalled a fact buried at any position with
100\% adherence, with no U-shaped dip. We then built a harder, size-swept test: a single agent must
retrieve the current payments contract, buried at $\sim$50\% depth among four hard-negative money
shapes, in a codebase inflated to a target token budget, against routing that delivered the one
relevant fact ($\sim$155 tokens). Accuracy did not degrade with dilution: both the frontier model
and the small one picked the current contract over the decoys at 10k, 50k, and 100k tokens (and the
frontier model up to $\sim$280k). A still-harder variant requiring a three-fact join with named
traps gave the same answer: no accuracy gap opened, even for the small model.

What is real, and large, is cost and capacity. Routing delivers the needed fact in $\sim$155 tokens
versus 10k--280k dumped into a single agent's prompt---68$\times$ to 1296$\times$ less context per
task---and the single-agent prompt that carries an entire microservices world ($\sim$280k tokens)
overflows the small model's 200k window outright (5/5 failures), while the per-agent routed prompt
always fits. So per-repo scoping plus routing is a cost-and-feasibility win, not an accuracy rescue;
we state it that way rather than the stronger claim we could not support.

One methods note belongs here, because it is the reason we trust the numbers above. An earlier run of
the decision-adherence test reported a 70\% versus 100\% effect on a frontier model. It was a scorer
artifact: a regular expression that negated on a substring false-flagged correct replies. Structured
output scoring plus raw-reply logging erased the effect, and we retract that number. The lesson---score
structured fields and log raw replies rather than pattern-match prose---is now baked into the harness,
and it is why the positive results in \Cref{sec:results} are scored on parsed artifacts (the actual
import line, the merged field) rather than on prose mentions.

\section{Discussion}
\label{sec:discussion}

The live contract result splits cleanly in two. The coordination \emph{signal}
helps across the tested capability tiers: without it, a frontier model, a small model, and a second
vendor's model all fail at 0.00, because the fact is absent from their context, not beyond their reasoning.
\emph{Acting} on the signal is partly capability-dependent: given the same plan, the frontier models
adapt perfectly (1.00) and the small model most of the time (0.60), with the misses tracing to
weaker execution (a stray edit, an incomplete migration) rather than to missing information. That
split matters for anyone worried that better models will make coordination unnecessary. On the
evidence here the information half is not closed by scale; the execution half already benefits from it.

It also explains when reactive detection is enough and when it is not. In the deterministic timing
model, a warning lands in time on sequential dependencies and the detector matches the planner. Under
scripted concurrent contention the warning arrives after the edit, so the detector tracks
no-coordination and the gap to the planner grows with $N$. In the live study, where timing and
information specificity change together, the evidence supports the combined proactive intervention
rather than scheduling alone.

Verification is easiest where ground truth is unambiguous---formal proofs, competitive programming,
passing test suites. Multi-agent code integration is one more such place: the merged tree either
builds and passes its acceptance check or it does not. NP-Bench exploits that to \emph{verify}
coordination end to end---it runs real agents in a real git integration and reads ground-truth
signals (clean-merge success, scope compliance, contract adaptation, cross-session decision
adherence) directly off the merged tree, rather than trusting each agent's self-report. Coordination
failures are invisible to single-agent verification: each agent, judged alone, did fine; the system
still broke on integration. The three-arm design (nothing / reactive / proactive) lets us measure
whether the coordination layer helped instead of asserting it, and the negative arm guards against
overclaiming.

\section{Threats to validity}
\label{sec:threats}

Tier A is a deterministic simulation of agent \emph{reaction};
it drives the real sensing, detection, routing, and planner, but the agent edits
themselves are scripted, so it validates mechanism and routing precision, not agent behavior. That
is what Tier B is for. The live cells span two model families (Claude frontier/Haiku and OpenAI
Codex), which guards against a single-vendor artifact, and the pivotal contract cell is run at
$K{=}15$, where the planner's Wilson interval $[0.80,1.00]$ separates cleanly from the baselines'
$[0.00,0.20]$ (Fisher's exact $p<10^{-4}$). The remaining live cells are $K{=}5$ pilots with wider
intervals, and the scenarios are hand-constructed; larger $K$ across every cell and a broader
scenario set are the obvious next runs.

The most important limitation is scope inference. In these experiments the declared scopes are
\emph{hand-seeded} in the harness (each item's stated files and consumed contracts), not inferred
from natural-language tasks. The planner's guarantee is only as good as those scopes, and inferring
them automatically (from a symbol graph or the service graph) is unbuilt work, not a solved
component; we claim the scheduling result, not automatic scope inference. Scope-leakage is also
measured on the simple shared-file case; leakage could be higher on subtler scopes. Finally, the
\arm{C1-detect} warning is deliberately vague. We argue this is realistic rather than a straw man---a
just-in-time ``someone may be touching $X$'' is what reactive systems actually emit---but a
specificity-versus-timing ablation (a detailed but late warning) would sharpen the comparison and is
future work.

\section{Conclusion}
\label{sec:conclusion}

Parallel coding agents collide at integration, and that collision is a scheduling problem. Given declared
scopes, scheduling around them (disjoint scopes for concurrent work, producers merged
before their consumers) turns conflict cleanup into conflict prevention. On NP-Bench the planner lifts
deterministic clean-integration from 1/9 to 9/9 and, with live agents from two vendors, rescues a
breaking contract change that no-coordination and reactive detection both miss (0.00/0.00$\rightarrow$1.00
on the frontier models, 0.60 on a small one), while agents respect their assigned scopes. A
cross-session memory closes a repeated-mistake gap identically on strong and weak models. Routing
facts to agents, which we expected to rescue long-context accuracy, does not at window-fitting
scales; its value is cost and capacity. The benefit held across the models we tested because it comes
from how work is allocated, not from model reasoning. What remains open is inferring scopes
automatically rather than hand-seeding them, re-planning as work changes, and moving from advisory
scopes to enforced ones.

\paragraph{Reproducibility.} \ifanon
\else The system (\sysname v0.17.0) and the NP-Bench harness are
open source at \mbox{\url{https://github.com/sumanyumuku98/Nerveplane}}.\fi{} Tier A is deterministic and CI-gated; Tier B is a set of live-agent runs that are
nondeterministic and reported with Wilson 95\% confidence intervals.

\bibliographystyle{plainnat}
\bibliography{references}

\begin{thebibliography}{19}
\providecommand{\natexlab}[1]{#1}
\providecommand{\url}[1]{\texttt{#1}}
\expandafter\ifx\csname urlstyle\endcsname\relax
  \providecommand{\doi}[1]{doi: #1}\else
  \providecommand{\doi}{doi: \begingroup \urlstyle{rm}\Url}\fi

\bibitem[{Anthropic}(2024)]{anthropic2024mcp}
{Anthropic}.
\newblock Introducing the model context protocol.
\newblock \url{https://www.anthropic.com/news/model-context-protocol}, 2024.
\newblock Accessed 2026-08-20.

\bibitem[Brooks(1975)]{brooks1975mythical}
Frederick~P. Brooks.
\newblock \emph{The Mythical Man-Month: Essays on Software Engineering}.
\newblock Addison-Wesley, 1975.

\bibitem[Brun et~al.(2011)Brun, Holmes, Ernst, and Notkin]{brun2011proactive}
Yuriy Brun, Reid Holmes, Michael~D. Ernst, and David Notkin.
\newblock Proactive detection of collaboration conflicts.
\newblock In \emph{ACM SIGSOFT FSE}, pages 168--178, 2011.

\bibitem[Chen et~al.(2021)Chen, Tworek, Jun, Yuan, Pinto, Kaplan,
  et~al.]{chen2021codex}
Mark Chen, Jerry Tworek, Heewoo Jun, Qiming Yuan, Henrique Ponde de~Oliveira
  Pinto, Jared Kaplan, et~al.
\newblock Evaluating large language models trained on code.
\newblock \emph{arXiv preprint arXiv:2107.03374}, 2021.

\bibitem[Chen et~al.(2024)Chen, Su, Zuo, Yang, Yuan, Chan, Yu, Lu, Hung, Qian,
  Qin, Cong, Xie, Liu, Sun, and Zhou]{chen2023agentverse}
Weize Chen, Yusheng Su, Jingwei Zuo, Cheng Yang, Chenfei Yuan, Chi-Min Chan,
  Heyang Yu, Yaxi Lu, Yi-Hsin Hung, Chen Qian, Yujia Qin, Xin Cong, Ruobing
  Xie, Zhiyuan Liu, Maosong Sun, and Jie Zhou.
\newblock {AgentVerse}: Facilitating multi-agent collaboration and exploring
  emergent behaviors.
\newblock \emph{ICLR}, 2024.

\bibitem[Conway(1968)]{conway1968committees}
Melvin~E. Conway.
\newblock How do committees invent?
\newblock \emph{Datamation}, 14\penalty0 (4):\penalty0 28--31, 1968.

\bibitem[Garey and Johnson(1979)]{garey1979computers}
Michael~R. Garey and David~S. Johnson.
\newblock \emph{Computers and Intractability: A Guide to the Theory of
  {NP}-Completeness}.
\newblock W. H. Freeman, 1979.

\bibitem[{Google}(2025)]{google2025a2a}
{Google}.
\newblock Announcing the {Agent2Agent} protocol ({A2A}).
\newblock
  \url{https://developers.googleblog.com/en/a2a-a-new-era-of-agent-interoperability/},
  2025.
\newblock Accessed 2026-08-20.

\bibitem[Hong et~al.(2024)Hong, Zhuge, Chen, Zheng, Cheng, Zhang, Wang, Wang,
  Yau, Lin, Zhou, Ran, Xiao, Wu, and Schmidhuber]{hong2024metagpt}
Sirui Hong, Mingchen Zhuge, Jonathan Chen, Xiawu Zheng, Yuheng Cheng, Ceyao
  Zhang, Jinlin Wang, Zili Wang, Steven Ka~Shing Yau, Zijuan Lin, Liyang Zhou,
  Chenyu Ran, Lingfeng Xiao, Chenglin Wu, and J{\"u}rgen Schmidhuber.
\newblock {MetaGPT}: Meta programming for a multi-agent collaborative
  framework.
\newblock \emph{ICLR}, 2024.

\bibitem[Jain et~al.(2025)Jain, Han, Gu, Li, Yan, Zhang, Wang, Solar-Lezama,
  Sen, and Stoica]{jain2024livecodebench}
Naman Jain, King Han, Alex Gu, Wen-Ding Li, Fanjia Yan, Tianjun Zhang, Sida
  Wang, Armando Solar-Lezama, Koushik Sen, and Ion Stoica.
\newblock {LiveCodeBench}: Holistic and contamination free evaluation of large
  language models for code.
\newblock \emph{ICLR}, 2025.

\bibitem[Jimenez et~al.(2024)Jimenez, Yang, Wettig, Yao, Pei, Press, and
  Narasimhan]{jimenez2024swebench}
Carlos~E. Jimenez, John Yang, Alexander Wettig, Shunyu Yao, Kexin Pei, Ofir
  Press, and Karthik Narasimhan.
\newblock {SWE-bench}: Can language models resolve real-world {GitHub} issues?
\newblock \emph{ICLR}, 2024.

\bibitem[Kahn(1962)]{kahn1962topological}
Arthur~B. Kahn.
\newblock Topological sorting of large networks.
\newblock \emph{Communications of the ACM}, 5\penalty0 (11):\penalty0 558--562,
  1962.

\bibitem[Li et~al.(2023)Li, Hammoud, Itani, Khizbullin, and
  Ghanem]{li2023camel}
Guohao Li, Hasan Abed Al~Kader Hammoud, Hani Itani, Dmitrii Khizbullin, and
  Bernard Ghanem.
\newblock {CAMEL}: Communicative agents for ``mind'' exploration of large
  language model society.
\newblock \emph{NeurIPS}, 2023.

\bibitem[Liu et~al.(2024)Liu, Lin, Hewitt, Paranjape, Bevilacqua, Petroni, and
  Liang]{liu2024lost}
Nelson~F. Liu, Kevin Lin, John Hewitt, Ashwin Paranjape, Michele Bevilacqua,
  Fabio Petroni, and Percy Liang.
\newblock Lost in the middle: How language models use long contexts.
\newblock \emph{Transactions of the ACL (TACL)}, 12:\penalty0 157--173, 2024.

\bibitem[Packer et~al.(2023)Packer, Wooders, Lin, Fang, Patil, Stoica, and
  Gonzalez]{packer2023memgpt}
Charles Packer, Sarah Wooders, Kevin Lin, Vivian Fang, Shishir~G. Patil, Ion
  Stoica, and Joseph~E. Gonzalez.
\newblock {MemGPT}: Towards {LLMs} as operating systems.
\newblock \emph{arXiv preprint arXiv:2310.08560}, 2023.

\bibitem[Park et~al.(2023)Park, O'Brien, Cai, Morris, Liang, and
  Bernstein]{park2023generative}
Joon~Sung Park, Joseph~C. O'Brien, Carrie~J. Cai, Meredith~Ringel Morris, Percy
  Liang, and Michael~S. Bernstein.
\newblock Generative agents: Interactive simulacra of human behavior.
\newblock In \emph{ACM UIST}, 2023.

\bibitem[Qian et~al.(2024)Qian, Liu, Liu, Chen, Dang, Li, Yang, Chen, Su, Cong,
  Xu, Li, Liu, and Sun]{qian2024chatdev}
Chen Qian, Wei Liu, Hongzhang Liu, Nuo Chen, Yufan Dang, Jiahao Li, Cheng Yang,
  Weize Chen, Yusheng Su, Xin Cong, Juyuan Xu, Dahai Li, Zhiyuan Liu, and
  Maosong Sun.
\newblock {ChatDev}: Communicative agents for software development.
\newblock \emph{ACL}, 2024.

\bibitem[Wu et~al.(2023)Wu, Bansal, Zhang, Wu, Li, Zhu, Jiang, Zhang, Zhang,
  Liu, Awadallah, White, Burger, and Wang]{wu2023autogen}
Qingyun Wu, Gagan Bansal, Jieyu Zhang, Yiran Wu, Beibin Li, Erkang Zhu,
  Li~Jiang, Xiaoyun Zhang, Shaokun Zhang, Jiale Liu, Ahmed~H. Awadallah,
  Ryen~W. White, Doug Burger, and Chi Wang.
\newblock {AutoGen}: Enabling next-gen {LLM} applications via multi-agent
  conversation.
\newblock \emph{arXiv preprint arXiv:2308.08155}, 2023.

\bibitem[Yang et~al.(2024)Yang, Jimenez, Wettig, Lieret, Yao, Narasimhan, and
  Press]{yang2024sweagent}
John Yang, Carlos~E. Jimenez, Alexander Wettig, Kilian Lieret, Shunyu Yao,
  Karthik Narasimhan, and Ofir Press.
\newblock {SWE-agent}: Agent-computer interfaces enable automated software
  engineering.
\newblock In \emph{NeurIPS}, 2024.

\end{thebibliography}

\end{document}